\documentclass[11pt]{article}

\usepackage[utf8]{inputenc}
\usepackage[T1]{fontenc}
\usepackage[margin=1in]{geometry}
\usepackage{microtype}
\usepackage{amsmath,amssymb,amsthm}
\usepackage{graphicx}
\usepackage{booktabs}
\usepackage{array}
\usepackage{multirow}
\usepackage{enumitem}
\usepackage{xcolor}
\usepackage[colorlinks=true,linkcolor=blue!50!black,citecolor=blue!50!black,urlcolor=blue!60!black]{hyperref}
\usepackage{url}
\usepackage[square,numbers,sort&compress]{natbib}
\usepackage{caption}
\usepackage{tikz}
\usetikzlibrary{arrows.meta,positioning,shapes.geometric,fit,calc}

\definecolor{srcBlue}{HTML}{1F4E79}
\definecolor{hashGreen}{HTML}{2E7D32}
\definecolor{cleanOrange}{HTML}{E65100}
\definecolor{lidPurple}{HTML}{6A1B9A}
\definecolor{dupTeal}{HTML}{00695C}
\definecolor{qualRed}{HTML}{C62828}
\definecolor{releaseGold}{HTML}{8D6E00}
\definecolor{softGray}{HTML}{555555}

\newtheorem{definition}{Definition}

\title{SomaliWeb v1: A Quality-Filtered Somali Web Corpus with a Matched Tokenizer and a Public Language-Identification Benchmark}

\author{
  Khalid Yusuf Dahir \\
  Independent researcher \\
  \texttt{khaliddahir0200@gmail.com} \\
}

\date{May 2026}

\begin{document}
\maketitle

\begin{abstract}
Somali is a Cushitic language of the Horn of Africa with $\sim$25 million speakers, yet no documented dedicated Somali pretraining corpus with a companion tokenizer and language-identification benchmark has been publicly released. Existing Somali text appears either inside multilingual distributions (HPLT v2, CC100, MADLAD-400, OSCAR, mC4) or in small, undocumented Somali-only uploads on Hugging Face (\S\ref{sec:related}). We introduce \textbf{SomaliWeb v1}, a quality-filtered Somali corpus of 819{,}322 documents ($\approx$303M tokens) built from three upstream sources (HPLT v2, CC100, Somali Wikipedia) through a six-stage reproducible pipeline. We release (i) the corpus, (ii) a matched BPE-16K tokenizer, and (iii) the first public side-by-side Somali benchmark of three production language identifiers. Our measurements reveal concrete quality defects in existing distributions: HPLT v2's ``cleaned'' Somali release retains \textbf{17.3\%} byte-exact duplicates, \textbf{56.1\%} of its documents contain fixable mojibake, and \textbf{10.7\%} of its byte-unique documents are near-duplicates at Jaccard $\tau=0.80$. Our BPE-16K tokenizer emits \textbf{40.2\% fewer tokens} than GPT-4's \texttt{cl100k\_base} on FLORES-200 Somali devtest as a tokenizer-level measurement; downstream language-model perplexity comparisons are deferred to a follow-up release.

\medskip
\noindent\textbf{Code:} \url{https://github.com/khaledyusuf44/somali-corpus}\\
\textbf{Dataset:} \url{https://huggingface.co/datasets/khaledyusuf44/somaliweb-v1}\\
\textbf{License:} Pipeline code MIT; corpus CC-BY-SA 4.0; this paper CC-BY 4.0.
\end{abstract}

\paragraph{Contributions.}
\begin{itemize}[leftmargin=1.5em,itemsep=2pt,topsep=2pt]
  \item \textbf{C1 (Artifact).} The first versioned, documented, single-language Somali pretraining corpus released with a companion tokenizer and language-identification benchmark: 819{,}322 documents, $\sim$303M whitespace-approximated tokens, with a 95/5 train/validation split, full dataset card, reproducibility manifest, and CC-BY-SA 4.0 license. Two prior Somali-tagged Hugging Face datasets (\citet{ibraahim2026fineweb} and \citet{farmerline2024somali}) are smaller, either undocumented or single-source, and (in the case of \citet{farmerline2024somali}) audio-plus-transcription rather than pretraining text (\S\ref{sec:somali-hf-datasets}).
  \item \textbf{C2 (Benchmark).} The first public Somali-specific benchmark of three widely-used language identifiers (\texttt{langdetect}, GlotLID v3, fastText \texttt{lid.176}) with per-class precision, recall, F1, throughput, and 95\% bootstrap confidence intervals.
  \item \textbf{C3 (Measurement).} Three concrete, quantified quality defects in HPLT v2's ``cleaned'' Somali distribution (17.3\% byte-duplicates, 56.1\% mojibake-bearing documents, 10.7\% near-duplicates) with per-phase retention and per-source breakdowns.
  \item \textbf{C4 (Tool).} A BPE-16K tokenizer trained on our cleaned corpus that is 40.2\% more token-efficient than GPT-4's \texttt{cl100k\_base} on FLORES-200 Somali devtest \emph{at the tokenizer-fertility level}, and ties HPLT-raw tokenizer fertility with a 30\% smaller training corpus. We frame this as a measurement of representational compression; downstream impact on language-model perplexity is left to future work.
\end{itemize}

\section{Introduction}
\label{sec:intro}

The ``tokenization tax'' paid by general-purpose language models on low-resource languages is well documented \citep{petrov2023tokenizers,ali2024tokenizer}. Somali, a Cushitic language of the Horn of Africa with Latin-script orthography and roughly 25 million speakers, is an exemplar of the mismatch: a major world language with active news, diaspora, and social-media ecosystems, yet with no standalone pretraining corpus released on Hugging Face or any other public registry. Somali text appears inside multilingual distributions (HPLT v2, CC100, MADLAD-400, OSCAR, mC4), but always as ``one of 100+ languages,'' without a dedicated release, dataset card, or documented construction pipeline.

This infrastructure gap has two downstream consequences. First, \textbf{practitioners cannot audit} what ``Somali training data'' means in any specific model: there is no canonical Somali corpus to measure against. Second, \textbf{researchers cannot iterate}: every Somali modeling effort must re-derive its own corpus from raw multilingual dumps, re-discovering the same quality issues each time.

We address both by releasing \textbf{SomaliWeb v1}: a corpus, a tokenizer, and a benchmark, with a documented six-stage pipeline that makes every filter auditable.

\begin{figure*}[t]
  \centering
  \resizebox{0.95\textwidth}{!}{%
  \begin{tikzpicture}[
    font=\sffamily\footnotesize,
    node distance=6mm and 10mm,
    src/.style   = {draw, rounded corners=2pt, fill=srcBlue!10,    text=srcBlue,     minimum width=34mm, minimum height=9mm,  align=center, inner sep=2pt},
    phase/.style = {draw, rounded corners=2pt, fill=#1!10,         text=#1,          minimum width=46mm, minimum height=12mm, align=center, line width=0.5pt, inner sep=2pt},
    rel/.style   = {draw, rounded corners=2pt, fill=releaseGold!15,text=releaseGold, minimum width=46mm, minimum height=12mm, align=center, line width=0.8pt},
    arrow/.style = {-{Stealth[length=2.2mm]}, thick, softGray},
    stat/.style  = {font=\sffamily\tiny, text=softGray, align=center}
  ]
    \node[src] (s1) {HPLT v2 \texttt{som\_Latn}\\[-1pt]\scriptsize 966{,}507 docs $\cdot$ 505M tok};
    \node[src, below=of s1] (s2) {CC100 \texttt{so}\\[-1pt]\scriptsize 396{,}524 docs $\cdot$ 81M tok};
    \node[src, below=of s2] (s3) {Wikipedia-so\\[-1pt]\scriptsize 9{,}021 docs $\cdot$ 2.5M tok};
    \node[phase=srcBlue, right=16mm of s2] (agg) {\textbf{Aggregate}\\[-1pt]\scriptsize 1{,}372{,}052 docs $\cdot$ 588M tok};
    \draw[arrow] (s1.east) -- (agg.west);
    \draw[arrow] (s2.east) -- (agg.west);
    \draw[arrow] (s3.east) -- (agg.west);
    \node[phase=hashGreen,   right=of agg] (p1) {\textbf{Phase 1}\\Byte-exact dedup\\[-1pt]\scriptsize SHA-256};
    \node[phase=cleanOrange, right=of p1]  (p2) {\textbf{Phase 2}\\ftfy + NFC + $\geq$50w\\[-1pt]\scriptsize mojibake repair};
    \node[phase=lidPurple,   right=of p2]  (p3) {\textbf{Phase 3}\\LID verify\\[-1pt]\scriptsize langdetect $\geq$0.50};
    \node[phase=dupTeal,  below=14mm of p3] (p4) {\textbf{Phase 4}\\MinHash near-dedup\\[-1pt]\scriptsize $(b,r){=}(16,4)$, $\tau{=}0.80$};
    \node[phase=qualRed,  left=of p4] (p5)  {\textbf{Phase 5}\\Char-5-gram quality\\[-1pt]\scriptsize drop bottom 15\%};
    \node[rel, left=of p5] (rel)            {\textbf{Phase 6 -- Release}\\Shuffle + 95/5 + BPE-16K\\[-1pt]\scriptsize \textbf{819{,}322 docs $\cdot$ 303M tok}};
    \draw[arrow] (agg.east) -- (p1.west);
    \draw[arrow] (p1.east)  -- (p2.west);
    \draw[arrow] (p2.east)  -- (p3.west);
    \draw[arrow] (p3.south) to[out=-90,in=90] (p4.north);
    \draw[arrow] (p4.west)  -- (p5.east);
    \draw[arrow] (p5.west)  -- (rel.east);
    \node[stat, above=1pt of p1] {$\downarrow$ 13.83\% $\cdot$ 1{,}182{,}360};
    \node[stat, above=1pt of p2] {$\downarrow$ 8.65\% $\cdot$ 1{,}080{,}040};
    \node[stat, above=1pt of p3] {$\downarrow$ 0.21\% $\cdot$ 1{,}077{,}804};
    \node[stat, below=1pt of p4] {$\downarrow$ 10.57\% $\cdot$ 963{,}908};
    \node[stat, below=1pt of p5] {$\downarrow$ 15.00\% $\cdot$ 819{,}322};
  \end{tikzpicture}%
  }
  \caption{SomaliWeb v1 --- the six-stage corpus construction pipeline with per-phase retention. See \S\ref{sec:method} for equations and \S\ref{sec:results} for retention tables.}
  \label{fig:pipeline}
\end{figure*}
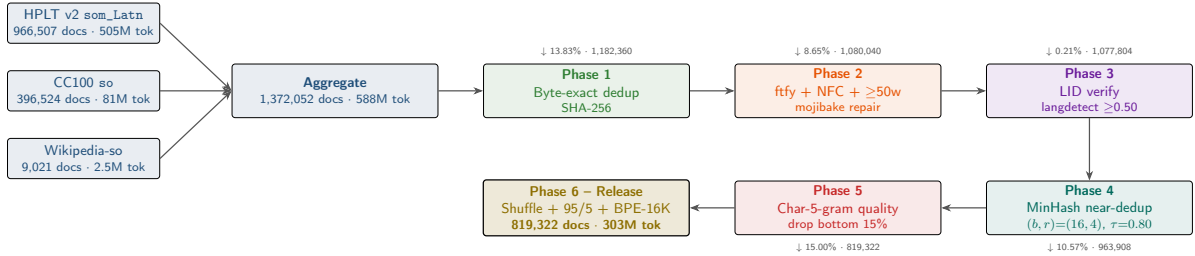

\paragraph{Roadmap.} \S\ref{sec:background} surveys the Horn of Africa language ecosystem and catalogs the current state of Somali resources. \S\ref{sec:related} reviews related work along three axes: web-scale multilingual corpora, low-resource African NLP, and language identification. \S\ref{sec:problem} formalizes the corpus construction problem. \S\ref{sec:method} describes our six-stage pipeline with per-phase equations. \S\ref{sec:setup} specifies experimental setup. \S\ref{sec:results} reports results. \S\ref{sec:analysis} analyzes errors and dialect coverage. \S\ref{sec:limits} lists limitations. \S\ref{sec:conclusion} concludes.

\section{Background: Languages of the Horn of Africa}
\label{sec:background}

The Horn of Africa (Somalia, Ethiopia, Eritrea, Djibouti, and Somali-speaking regions of Kenya) hosts roughly 130 million speakers across Afroasiatic (Cushitic, Semitic, Omotic) and Nilo-Saharan families. Five languages dominate by speaker count; four are critically under-resourced in NLP. Table~\ref{tab:horn} summarizes the resource landscape.

\begin{table}[t]
\centering
\small
\caption{Horn of Africa languages: speakers and corpus availability. Speaker counts from Ethnologue (2024). ``Dedicated corpus'' means a stand-alone, publicly released, versioned Somali-only dataset with a curation pipeline; SomaliWeb v1 is the first such release for any Horn of Africa language.}
\label{tab:horn}
\begin{tabular}{lclllccc>{\centering\arraybackslash}p{1.3cm}}
\toprule
Language & ISO & Spk. (M) & Script & Family & HPLT v2 & CC100 & MADLAD & Dedicated\\
\midrule
Amharic  & amh & 57 & Ge'ez & Semitic  & \checkmark & \checkmark & \checkmark & ---\\
Oromo    & orm & 37 & Latin & Cushitic & \checkmark & \checkmark & \checkmark & ---\\
Somali   & som & 25 & Latin & Cushitic & \checkmark & \checkmark & \checkmark & \textbf{ours} \\
Tigrinya & tir &  9 & Ge'ez & Semitic  & \checkmark & \checkmark & \checkmark & ---\\
Afar     & aar &  2 & Lat./Ge'ez & Cushitic & --- & --- & --- & ---\\
\bottomrule
\end{tabular}
\end{table}

\paragraph{Observations.}

\begin{enumerate}[leftmargin=1.5em,itemsep=2pt,topsep=2pt]
  \item \textbf{Population growth, resource stagnation.} The five languages in Table~\ref{tab:horn} collectively represent $\sim$130M first-language speakers, of whom Somali alone accounts for 25M, comparable to the combined speaker counts of Dutch (24M) and Swedish (10M). Yet Dutch and Swedish each have numerous dedicated corpora (SoNaR, OSCAR-nl, Europarl-nl; SIC, KB-corpus) of sizes far exceeding what is available for any Horn of Africa language.
  \item \textbf{Inclusion $\neq$ documentation.} Four of the five Horn languages appear in every major multilingual distribution, yet none has a stand-alone, documented, versioned release. ``Included'' is not ``usable'': a practitioner who wants ``Somali training data'' currently has no canonical reference artifact.
  \item \textbf{Ready-made multilingual corpora are not drop-in usable.} Our Phase 1--4 measurements (\S\ref{sec:results}) show that HPLT v2's Somali partition (the largest single upstream Somali source) contains 17.3\% byte-exact duplicates and 56.1\% mojibake-bearing documents after HPLT's own cleaning pass. ``Available'' and ``clean'' are distinct claims; the former does not imply the latter.
\end{enumerate}

\paragraph{Why Somali first.} We focus SomaliWeb v1 on Somali because (i) Somali has the largest extant upstream footprint among Horn languages (966K HPLT v2 documents, $\sim$505M tokens), making it tractable as a corpus-engineering target; (ii) Somali's Latin orthography simplifies pipeline tooling versus Ge'ez-script Amharic/Tigrinya, which would require separate script-aware normalization; (iii) the first author is a Somali speaker, enabling qualitative audit of pipeline outputs. We explicitly flag Amharic and Oromo as natural successors to this work.

\section{Related Work}
\label{sec:related}

\subsection{Web-scale multilingual corpora}

Somali is present in the following major multilingual distributions:

\begin{itemize}[leftmargin=1.5em,itemsep=2pt,topsep=2pt]
  \item \textbf{CC100} \citep{wenzek2020ccnet,conneau2020xlmr}: 81 MB of Somali text derived from CommonCrawl via the CCNet pipeline. No per-language versioning; no dataset card; known length-tail issues.
  \item \textbf{OSCAR} (multiple releases: 23.01 \citep{abadji2022oscar}, 24.05): deduplicated CommonCrawl. The Somali partition is gated behind Hugging Face access and lacks standalone documentation.
  \item \textbf{mC4} \citep{xue2021mt5}: mT5 training data. The Somali partition exists but is distributed as a subset of 101 languages with no Somali-specific curation.
  \item \textbf{HPLT v2} \citep{burchell2025hplt}: a recent high-quality multilingual release. The Somali (\texttt{som\_Latn}) partition is 918 MB compressed, $\sim$505M tokens. Our measurements (\S\ref{sec:results}) show that HPLT v2's ``cleaned'' Somali release retains 17.3\% byte-exact duplicates.
  \item \textbf{MADLAD-400} \citep{kudugunta2023madlad}: an audited multilingual dataset; includes a Somali partition without standalone release.
  \item \textbf{CulturaX} \citep{nguyen2024culturax} and \textbf{FineWeb-2} \citep{penedo2025fineweb2}: recent frontier corpora. Somali content is present but not separately cataloged.
\end{itemize}

\paragraph{Gap.} None of the above releases a Somali-only dataset with a documented construction pipeline, per-phase audit, or companion tokenizer. SomaliWeb v1 is the first such release.

\subsection{Low-resource African NLP}

Masakhane has catalyzed African-language NLP through named-entity \citep{adelani2021masakhaner}, news classification \citep{adelani2023masakhanews}, and machine-translation benchmarks. Pretrained models for African languages include AfriBERTa \citep{ogueji2021afriberta}, AfroLM \citep{dossou2022afrolm}, AfroXLMR, SERENGETI, and Glot500 \citep{imani2023glot500}. Somali is included in several of these as one of 100+ languages. However, \emph{all cited works use the existing multilingual corpora as-is}; none audit the Somali partition or re-derive a cleaned Somali dataset. Our work is complementary: we produce the data asset their successors can build on.

\subsection{Language identification and tokenization on low-resource languages}

Language identification (LID) is a prerequisite for any corpus construction pipeline. Production LID tools include Google's \texttt{cld3}, \texttt{langdetect} (a Java-to-Python port of Google's Java LID), fastText \texttt{lid.176} \citep{joulin2017fasttext}, GlotLID \citep{kargaran2023glotlid}, and OpenLID \citep{burchell2023openlid}. CommonLID \citep{commoncrawl2025commonlid} recently re-evaluated LID on web data but does not publish per-class Somali metrics. \emph{No prior work publishes side-by-side per-class F1 for the three dominant Somali-covering LID tools on a Somali-specific test set.} We fill this gap in \S\ref{sec:results}.

On tokenization, \citet{petrov2023tokenizers} quantify the unfairness introduced by English-centric tokenizers; \citet{ali2024tokenizer} measure the ``tokenizer tax'' as a dollar-cost multiplier on commercial APIs. We extend both lines by publishing the first concrete Somali fertility comparison between a native corpus-matched tokenizer and GPT-4's \texttt{cl100k\_base}.

\subsection{Existing Somali datasets on Hugging Face}
\label{sec:somali-hf-datasets}

A keyword search of the Hugging Face Hub for ``somali'' returns roughly 250 datasets at time of writing (April 2026). Almost all fall outside the pretraining-text category we target: 68~hours of automatic-speech-recognition audio \citep{ddd2026somaliasr}; bilingual sentence-pairs for machine translation; Alpaca-style instruction-tuning data translated into Somali; and large multilingual collections (BibleNLP, FLORES-Plus, mC4) that include a Somali subset without dedicated curation.

The two Somali-tagged HF artifacts closest to our framing are:

\begin{itemize}[leftmargin=1.5em,itemsep=2pt,topsep=2pt]
  \item \citet{ibraahim2026fineweb} (\texttt{IbraahimLab/fineweb-somali}, Feb 2026): a single-source scrape of BBC~Somali news articles (collected January~2026). \textbf{Size:} 4{,}910 documents, ${\sim}37$\,MB on disk (HF size category \texttt{1K<n<10K}). MIT license, restricted to ``educational and research purposes only.'' No companion artifacts; no documented construction pipeline beyond the dataset card. Notwithstanding its name, the artifact is not derived from FineWeb-2 \citep{penedo2025fineweb2}.
  \item \citet{farmerline2024somali} (\texttt{FarmerlineML/somali\_cleaned\_dataset}, June 2024): 2{,}936 audio-plus-transcription rows ($\approx$2.79\,GB total, dominated by the audio modality). \textbf{No declared license; the \texttt{README.md} is empty.} The schema (audio file + transcription text) implies an ASR-style use case, but no documentation states this explicitly. We list it here for completeness; despite its name, the artifact is not a pretraining-text corpus.
\end{itemize}

Table~\ref{tab:somali-comparators} compares both against SomaliWeb~v1.

\begin{table}[h]
\centering
\small
\caption{Somali-only artifacts on Hugging Face closest to our framing. SomaliWeb v1 is the only artifact with a documented construction pipeline, declared license, schema documentation, and companion tokenizer + LID benchmark. \textsuperscript{$\dagger$}\texttt{FarmerlineML/somali\_cleaned\_dataset} is included for completeness but is an audio-plus-transcription dataset rather than pretraining text.}
\label{tab:somali-comparators}
\begin{tabular}{lrrll}
\toprule
Dataset & Size & Docs & License & Companion artifacts \\
\midrule
\texttt{IbraahimLab/fineweb-somali}    & 37\,MB  & 4{,}910          & MIT  & none \\
\texttt{FarmerlineML/somali\_cleaned}\textsuperscript{$\dagger$}  & 2.79\,GB & 2{,}936       & ---  & none \\
\textbf{SomaliWeb v1 (ours)}           & 485\,MB & 819{,}322        & CC-BY-SA 4.0 & 2 tokenizers + LID benchmark + SHASUMS \\
\bottomrule
\end{tabular}
\end{table}

We do not claim SomaliWeb v1 is the only Somali-text artifact on the Hub. We do claim it is the only one (i)~with a documented six-stage construction pipeline whose intermediate statistics are reproducible, (ii)~accompanied by a matched tokenizer, (iii)~accompanied by a per-class Somali LID benchmark, and (iv)~at the 100K-1M document scale needed for tokenizer training and downstream language-model pretraining experiments.

\section{Problem Formulation}
\label{sec:problem}

We formalize Somali corpus construction as a constrained selection problem over the union of upstream sources.

\paragraph{Notation.} Let $\mathcal{S} = \{S_1, \ldots, S_n\}$ be a finite family of upstream document sets. Each $S_i$ is a multiset of documents $d \in \Sigma^*$ over the Unicode alphabet $\Sigma$. Each source has an unknown per-document quality distribution $Q_i$ and an unknown language distribution $L_i$ over language labels $\ell \in \mathcal{L}$.

\begin{definition}[Target-language corpus]
A target-language corpus $\mathcal{C}^{(\ell^*)} \subseteq \bigcup_i S_i$ is a subset of the union of upstream documents satisfying three properties:
\begin{enumerate}[leftmargin=2em,itemsep=2pt,topsep=2pt]
  \item[(i)]   \textbf{Language purity:} $\forall d \in \mathcal{C} : \mathrm{lang}(d) = \ell^*$ under some reference language identifier;
  \item[(ii)]  \textbf{Novelty:} $\forall d_i, d_j \in \mathcal{C}, i \neq j : J(d_i, d_j) < \tau$ where $J$ is Jaccard similarity on a chosen shingle space and $\tau \in [0, 1]$;
  \item[(iii)] \textbf{Quality:} $\forall d \in \mathcal{C} : q(d) \geq q_{\min}$ under a quality score $q : \Sigma^* \to [0, 1]$.
\end{enumerate}
\end{definition}

\begin{definition}[Pipeline composition]
A corpus construction pipeline is a sequence of filtering operators $F_1, \ldots, F_k$ each mapping a document multiset to a subset. The pipeline output is $\mathcal{C} = F_k \circ F_{k-1} \circ \cdots \circ F_1\!\left(\bigcup_i S_i\right)$.
\end{definition}

\paragraph{Objective.}
We select $\mathcal{C}$ to minimize a weighted composite loss:
\begin{equation}
  \mathcal{L}(\mathcal{C}) = \alpha\, F(\mathcal{T}_\mathcal{C}, \mathcal{D}_{\text{eval}}) + \beta\, R(\mathcal{C}) + \gamma\, N_{\neg \ell^*}(\mathcal{C})
  \label{eq:objective}
\end{equation}
subject to $|\mathcal{C}| \geq N_{\min}$, where $F(\mathcal{T}_\mathcal{C}, \mathcal{D}_{\text{eval}})$ is the fertility of a tokenizer trained on $\mathcal{C}$, $R(\mathcal{C})$ is within-corpus redundancy, $N_{\neg \ell^*}(\mathcal{C})$ is the number of non-target-language documents retained, and $N_{\min}$ is a lower bound below which tokenizer training is statistically unreliable.

\section{Methodology}
\label{sec:method}

Our pipeline comprises six phases executed sequentially. Each phase is specified by an input, an output, a filter equation, and a measurable retention rate. All hyper-parameters are stored in a single YAML config; all random seeds are fixed at 0.

\subsection{Source aggregation}

We aggregate three Somali sources:
\begin{itemize}[leftmargin=1.5em,itemsep=2pt,topsep=2pt]
  \item \textbf{HPLT v2} \texttt{som\_Latn} \citep{burchell2025hplt}: 918 MB compressed, 966{,}507 documents, $\approx$505M approx.\ tokens.
  \item \textbf{CC100} \texttt{so} \citep{wenzek2020ccnet}: 81 MB compressed, 396{,}524 documents, $\approx$81M approx.\ tokens.
  \item \textbf{Somali Wikipedia} 2023-11-01 dump, extracted via \texttt{wikiextractor}: 9{,}021 articles, $\approx$2.5M approx.\ tokens.
\end{itemize}
Raw union: \textbf{1{,}372{,}052 documents, $\approx$588M tokens}.

\subsection{Phase 1 --- Byte-exact deduplication}

We hash each document after lowercase normalization and whitespace collapse:
\begin{equation}
  \hat{h}(d) = \mathrm{SHA256}\!\left(\mathrm{collapse}_{\text{ws}}(\mathrm{lower}(d))\right)
\end{equation}
and keep the first occurrence of each distinct hash. Formally,
\begin{equation}
  \mathcal{C}_1 = \{d \in \textstyle\bigcup_i S_i : \hat{h}(d) \notin \hat{H}_{<}\}
\end{equation}
where $\hat{H}_{<}$ is the set of hashes seen prior to $d$ in the deterministic iteration order.

\textbf{Retention.} 1{,}182{,}360 / 1{,}372{,}052 = 86.17\%. Drop rate 13.83\%. Per-source byte-dup rate: HPLT v2 17.27\%, CC100 5.49\%, Wikipedia 0.16\%.

\subsection{Phase 2 --- Normalization and length filter}

We apply four operators in sequence: \texttt{ftfy} mojibake repair, Unicode NFC normalization, whitespace collapse, and repeated-character run collapse (\texttt{aaaaa} $\to$ \texttt{aaa}). Documents below 50 whitespace words are dropped.

\textbf{Retention.} 1{,}080{,}040 / 1{,}182{,}360 = 91.35\%. Drop rate 8.65\%, entirely from the length filter (102{,}320 short docs). Of surviving documents, \textbf{56.06\% had at least one character fixed by} \texttt{ftfy} \citep{speer2019ftfy} (447{,}736 / 798{,}624 HPLT v2 documents specifically).

\subsection{Phase 3 --- Language identification}

We run seeded \texttt{langdetect} on each document and retain those for which Somali is the top-1 language with confidence $\geq 0.50$:
\begin{equation}
  \mathcal{C}_3 = \big\{ d \in \mathcal{C}_2 : \arg\max_\ell P_{\text{ld}}(\ell \mid d) = \mathrm{som} \,\wedge\, P_{\text{ld}}(\mathrm{som} \mid d) \geq 0.5 \big\}
\end{equation}
A second pass with GlotLID v3 \citep{kargaran2023glotlid} tags dialect (\texttt{som\_Latn} vs.\ \texttt{ymm\_Latn}).

\textbf{Retention.} 1{,}077{,}804 / 1{,}080{,}040 = 99.79\%. The 0.21\% removed are primarily English-language leakage (1{,}743 documents, 77.8\% of drops).

\subsection{Phase 4 --- MinHash near-duplicate removal}

We apply LSH-accelerated MinHash \citep{broder1997resemblance,leskovec2020mining} with exact Jaccard verification.

\paragraph{Shingling.} Word-3-grams: $G(d) = \{w_i w_{i+1} w_{i+2} : i = 1, \ldots, |d|-2\}$.

\paragraph{Jaccard similarity.}
\begin{equation}
  J(A, B) = \frac{|A \cap B|}{|A \cup B|}
\end{equation}

\paragraph{MinHash estimator.} For $k=64$ independent hash functions $\{h_1, \ldots, h_k\}$, the MinHash signature of $d$ is $\sigma(d) = (\min_{g \in G(d)} h_1(g), \ldots, \min_{g \in G(d)} h_k(g))$. The estimator satisfies
\begin{equation}
  \mathbb{E}\!\left[\frac{1}{k}\sum_{i=1}^{k} \mathbb{1}[\sigma(d_A)_i = \sigma(d_B)_i]\right] = J(G(d_A), G(d_B))
\end{equation}
with standard error $\sigma_{\hat J} = \sqrt{J(1-J)/k}$; for $k=64$ and $J=0.5$ this is $\approx 0.0625$.

\begin{figure}[t]
  \centering
  \includegraphics[width=0.95\columnwidth]{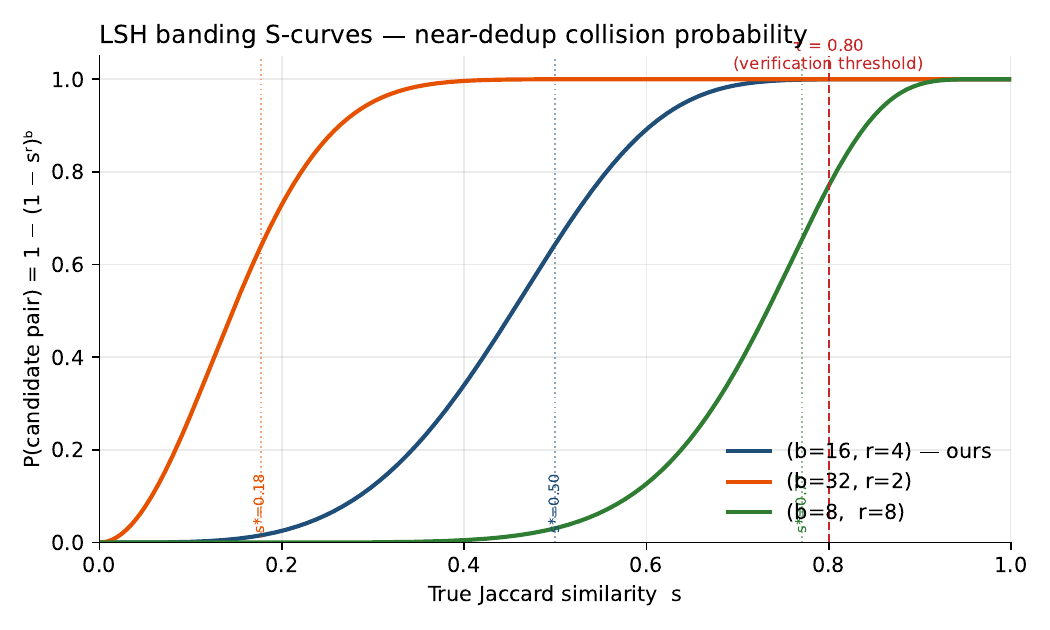}
  \caption{LSH S-curves~\eqref{eq:scurve} for three $(b,r)$ configurations. Our choice of $(16,4)$ is balanced around $s^* \approx 0.50$ with near-certain capture at $\tau = 0.80$.}
  \label{fig:scurve}
\end{figure}

\paragraph{LSH banding.} We partition the $k=64$-dimensional signatures into $b=16$ bands of $r=4$ rows each. Two documents are \emph{candidates} if they agree on at least one full band. The probability that two documents with true Jaccard $s$ are flagged candidates is
\begin{equation}
  P_{\text{cand}}(s; b, r) = 1 - (1 - s^r)^b.
  \label{eq:scurve}
\end{equation}
With $(b, r) = (16, 4)$, $P_{\text{cand}}(0.80) = 0.984$ and $P_{\text{cand}}(0.50) = 0.632$, giving $s^* = (1/b)^{1/r} \approx 0.50$. See Figure~\ref{fig:scurve}.

\paragraph{Verification.} For each candidate pair we compute exact Jaccard and retain the pair iff $J \geq \tau = 0.80$. Verified pairs are merged with union-find into clusters; within each cluster we keep the longest document (tie-broken by lexicographic ID).

\textbf{Retention.} 963{,}908 / 1{,}077{,}804 = 89.43\%. 82{,}679 clusters covering 196{,}575 documents; 113{,}896 documents removed.

\subsection{Phase 5 --- Character-n-gram quality filter}

We score each document by character-5-gram coverage against a clean Somali Wikipedia seed.

\paragraph{Seed.} All Wikipedia-so articles with $\geq$200 words: $|\mathcal{W}| = 2{,}221$ articles, $|G^{(5)}_{\text{seed}}| = 828{,}294$ distinct character-5-grams.

\paragraph{Coverage.}
\begin{equation}
  \mathrm{cov}(d) = \frac{|G^{(5)}(d) \cap G^{(5)}_{\text{seed}}|}{|G^{(5)}(d)|}
\end{equation}
We drop the bottom 15\% by coverage; empirically this corresponds to threshold $\mathrm{cov} \geq 0.9029$.

\textbf{Retention.} 819{,}322 / 963{,}908 = 85.00\%. Per-source drop rate reveals source-quality asymmetry: Wikipedia 20.61\%, HPLT 18.18\%, CC100 5.76\%.

\paragraph{Rationale.} No labeled Somali quality data exists. Set-coverage against a clean seed is the cheapest signal that correlates with ``looks like fluent Somali'' and requires zero negative examples. We discuss the trained-classifier alternative in \S\ref{sec:limits}.

\subsection{Phase 6 --- Release and tokenizer}

We shuffle (seed 0), split 95/5 train/val, and train a BPE-16K tokenizer on the train split. We evaluate fertility on FLORES-200 Somali devtest (1{,}012 sentences):
\begin{equation}
  F(\mathcal{T}, \mathcal{D}) = \frac{1}{|\mathcal{D}|} \sum_{s \in \mathcal{D}} \frac{|\mathcal{T}(s)|}{|\mathrm{words}(s)|}
\end{equation}

\section{Experimental Setup}
\label{sec:setup}

\paragraph{Hardware.} All experiments run on a single MacBook Pro M4 Pro (12 performance cores, 24 GB unified memory). No distributed compute.

\paragraph{Software.} Python 3.10. Pinned versions: \texttt{numpy<2}, \texttt{ftfy==6.1.3}, \texttt{langdetect==1.0.9}, \texttt{tokenizers==0.15.2}, \texttt{datasets==2.19.0}, \texttt{zstandard==0.22.0}, \texttt{tiktoken==0.7.0}. Full \texttt{requirements.txt} in Appendix~\ref{app:repro}.

\paragraph{Determinism.} All seeds fixed at 0: \texttt{random.seed(0)}, \texttt{np.random.seed(0)}, \texttt{DetectorFactory.seed = 0}, tokenizer trainer \texttt{shuffle\_seed=0}. With pinned versions the full pipeline is bit-exactly reproducible.

\paragraph{Wall-clock budget.}

\begin{table}[h]
\centering
\small
\begin{tabular}{lr}
\toprule
Phase & Elapsed \\
\midrule
1 -- merge + exact dedup        & 83 s  \\
2 -- clean + normalize          & 33 min \\
3 -- LID verify (10 workers)    & 14 min \\
4 -- MinHash near-dedup         & 5 min  \\
5 -- quality filter             & 5.3 min \\
6 -- structure + tokenizer      & $\sim$2 min \\
\midrule
\textbf{Total} & \textbf{$\sim$1 h} \\
\bottomrule
\end{tabular}
\end{table}

\section{Results}
\label{sec:results}

\subsection{Pipeline retention}

\begin{figure}[t]
  \centering
  \includegraphics[width=0.95\columnwidth]{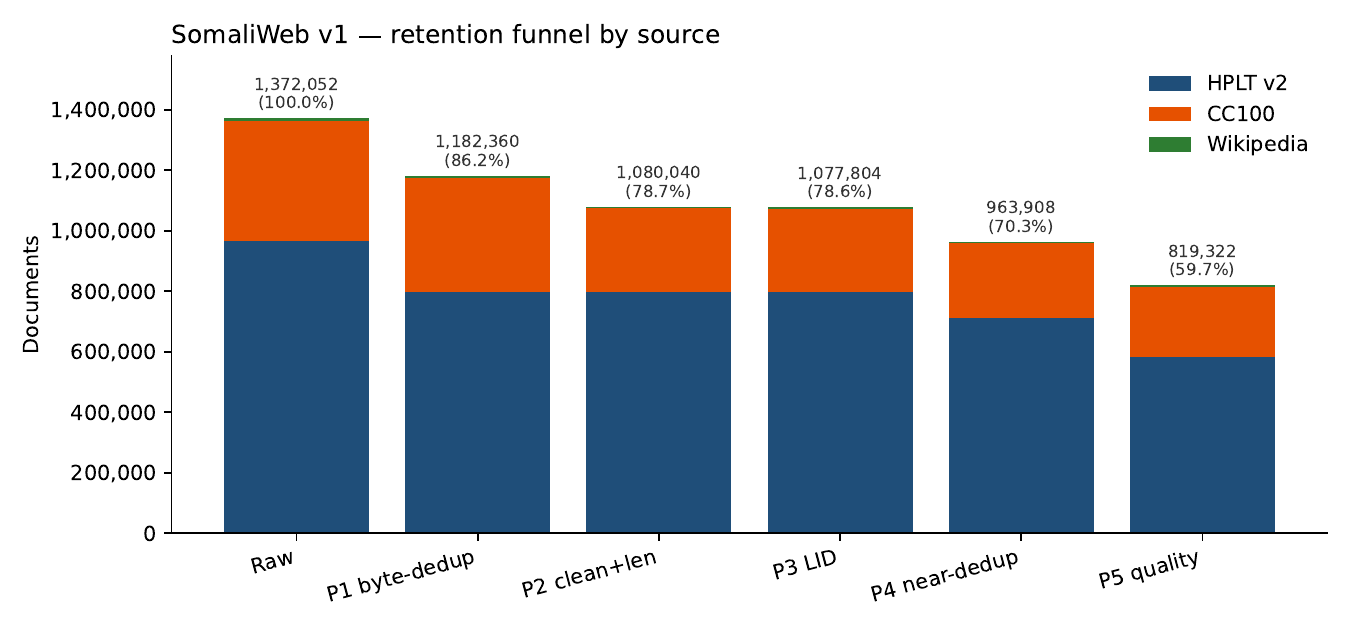}
  \caption{Per-source retention across the six pipeline phases.}
  \label{fig:funnel}
\end{figure}

\begin{table}[t]
\centering
\small
\caption{Pipeline retention. Final corpus is 59.7\% of the raw aggregated input.}
\label{tab:retention}
\begin{tabular}{lrrr}
\toprule
Stage & Documents & Retention & Drop \\
\midrule
Raw union                  & 1{,}372{,}052 & 100.00\% & --- \\
Phase 1 -- byte-dedup      & 1{,}182{,}360 & 86.17\%  & 13.83\% \\
Phase 2 -- clean + $\geq$50w & 1{,}080{,}040 & 78.72\%  & 8.65\%  \\
Phase 3 -- LID verify      & 1{,}077{,}804 & 78.56\%  & 0.21\%  \\
Phase 4 -- near-dedup      &   963{,}908 & 70.25\%  & 10.57\% \\
Phase 5 -- quality filter  & \textbf{819{,}322} & \textbf{59.72\%} & 15.00\% \\
\bottomrule
\end{tabular}
\end{table}

\subsection{Quality defects in HPLT v2 ``cleaned'' Somali}

\begin{table}[h]
\centering
\small
\caption{Three quantified quality defects in HPLT v2's ``cleaned'' \texttt{som\_Latn} distribution.}
\begin{tabular}{lr}
\toprule
Defect (within HPLT v2 input) & Rate \\
\midrule
Byte-exact duplicates (Phase 1)              & 17.27\% \\
\texttt{ftfy}-fixable mojibake (Phase 2)     & 56.06\% \\
Near-dups at $\tau=0.80$ among byte-uniques (Phase 4) & 10.67\% \\
\bottomrule
\end{tabular}
\end{table}

These three findings are concrete, auditable quality gaps in a widely-used ``cleaned'' distribution. Naïve consumption of HPLT v2 \texttt{som\_Latn} carries all three into downstream training.

\subsection{Somali language-identification benchmark}
\label{sec:lid}

Test set: 200 rows, 40 per language across $\{$\texttt{en}, \texttt{so}, \texttt{ar}, \texttt{fr}, \texttt{sw}$\}$, annotated by a Somali speaker. Evaluated models: \texttt{langdetect} (with \texttt{DetectorFactory.seed=0}), GlotLID v3, fastText \texttt{lid.176}.

\begin{figure*}[t]
  \centering
  \includegraphics[width=0.95\textwidth]{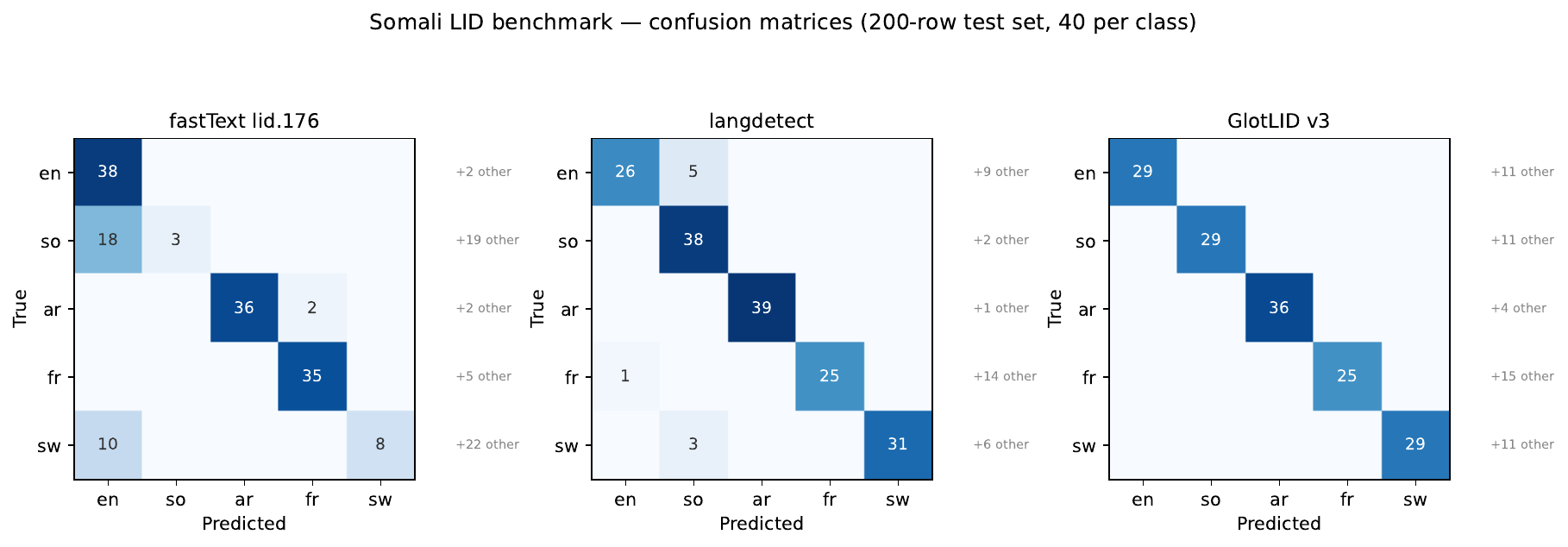}
  \caption{Somali LID confusion matrices on the 200-row test set (40 per class). \texttt{langdetect} dominates on Somali recall.}
  \label{fig:lid}
\end{figure*}

Point estimates with 95\% bootstrap confidence intervals (500 resamples, seed = 0):

\begin{table}[h]
\centering
\small
\caption{LID per-class metrics with 95\% bootstrap CIs.}
\label{tab:lid}
\begin{tabular}{lrrr}
\toprule
Model & Acc. & Docs/sec & \textbf{Somali F1 [95\% CI]} \\
\midrule
fastText \texttt{lid.176}    & 0.600 & 178{,}823 & 0.140 [0.000, 0.286] \\
GlotLID v3                   & 0.740 &  3{,}945 & 0.829 [0.712, 0.919] \\
\textbf{\texttt{langdetect}} & \textbf{0.795} & 417 & \textbf{0.884 [0.796, 0.945]} \\
\bottomrule
\end{tabular}
\end{table}

\paragraph{Principal finding.} \texttt{langdetect}, the oldest of the three (a Java-to-Python port of pre-deep-learning LID), achieves the highest Somali F1 on our test set, reversing the a priori assumption that newer fastText-based models would dominate. The CIs for \texttt{langdetect} and GlotLID v3 overlap, so we cannot claim statistical significance of the difference at $n = 40$ Somali rows; however, \texttt{langdetect}'s point estimate is higher and its lower CI bound (0.796) sits less than one percentage point below GlotLID v3's point estimate. fastText \texttt{lid.176}'s Somali recall is 0.075 with a CI lower bound of 0.000; its low TP count of 3 makes the F1 estimator unstable. GlotLID v3's 11 Somali misses go to unrelated Latin-script African languages (Fulfulde, Oromo, Kinyarwanda, Wolof, Bambara) rather than the defensible Maay Maay (\texttt{ymm\_Latn}) sister language. This pattern (confusing Somali with phylogenetically distant Latin-script African languages) makes GlotLID's ``not-Somali'' verdict unreliable as a single-stage filter for Somali-dedicated corpus construction.

\subsection{Tokenizer fertility on FLORES-200 Somali devtest}

\begin{figure}[t]
  \centering
  \includegraphics[width=0.95\columnwidth]{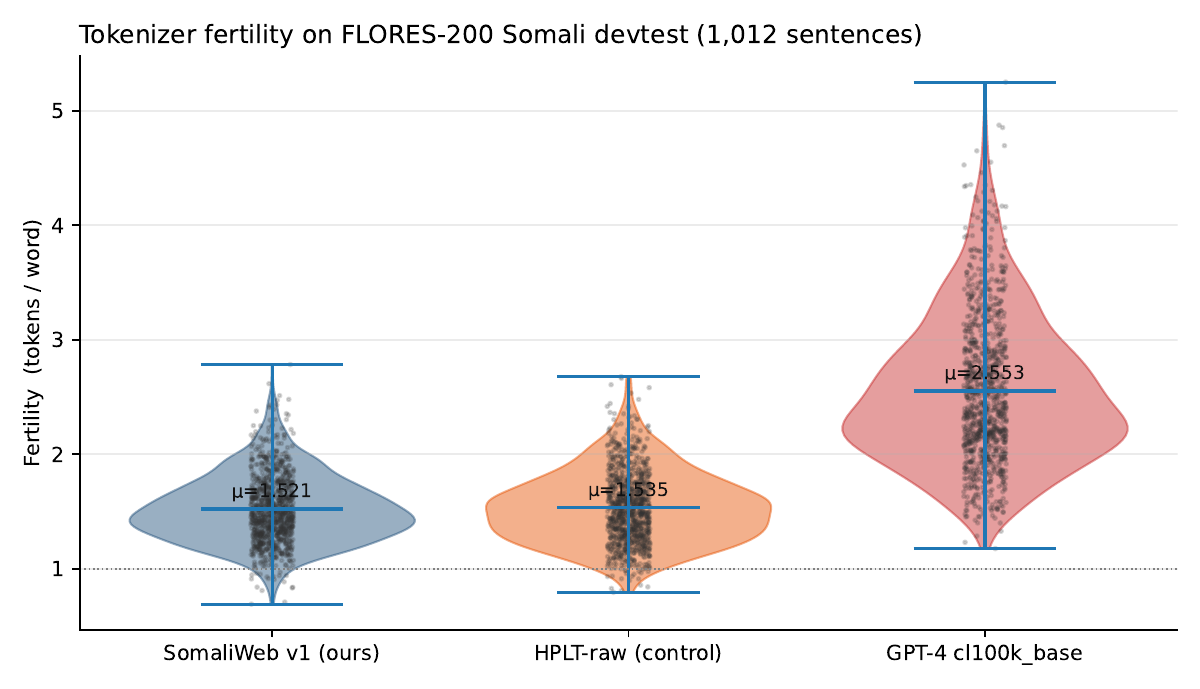}
  \caption{Tokenizer fertility distribution on FLORES-200 Somali devtest (1{,}012 sentences). SomaliWeb v1 ties HPLT-raw at 30\% smaller training corpus, and emits 40.2\% fewer tokens than GPT-4's \texttt{cl100k\_base}.}
  \label{fig:fertility}
\end{figure}

\begin{table}[h]
\centering
\small
\caption{Tokenizer fertility on 1{,}012 FLORES-200 Somali devtest sentences.}
\label{tab:fertility}
\begin{tabular}{lrrrr}
\toprule
Tokenizer & Vocab & Tokens & Words & \textbf{Fertility} $\downarrow$ \\
\midrule
\textbf{SomaliWeb v1} (ours)    & 16K   & 35{,}867 & 23{,}322 & \textbf{1.538} \\
HPLT-raw (control)              & 16K   & 35{,}854 & 23{,}322 & 1.537 \\
GPT-4 \texttt{cl100k\_base}     & 100K  & 60{,}010 & 23{,}322 & 2.573 \\
\bottomrule
\end{tabular}
\end{table}

\paragraph{Findings.}
\begin{enumerate}[leftmargin=1.5em,itemsep=2pt,topsep=2pt]
  \item \textbf{SomaliWeb v1 matches HPLT-raw tokenizer fertility with a 30\% smaller training corpus.} Tokenizer quality does not degrade under curation; it may improve per training-token information density.
  \item \textbf{SomaliWeb v1 is 40.2\% more token-efficient than GPT-4's \texttt{cl100k\_base}} \emph{at the tokenizer-fertility level}. On the same 1{,}012 FLORES sentences, \texttt{cl100k\_base} emits 60{,}010 tokens; SomaliWeb v1 emits 35{,}867 --- a 1.67$\times$ token-count multiplier. This is an upper bound on per-request inference-cost overhead, modulo model-specific batching and cache effects; the corresponding downstream perplexity comparison is left to future work.
\end{enumerate}

\subsection{Ablations (planned for camera-ready)}

We plan four ablations removing one pipeline phase at a time and measuring downstream tokenizer fertility, plus a tiny character-level LM trained on each ablated corpus reporting validation loss as a stronger downstream signal.

\begin{figure}[t]
  \centering
  \includegraphics[width=0.95\columnwidth]{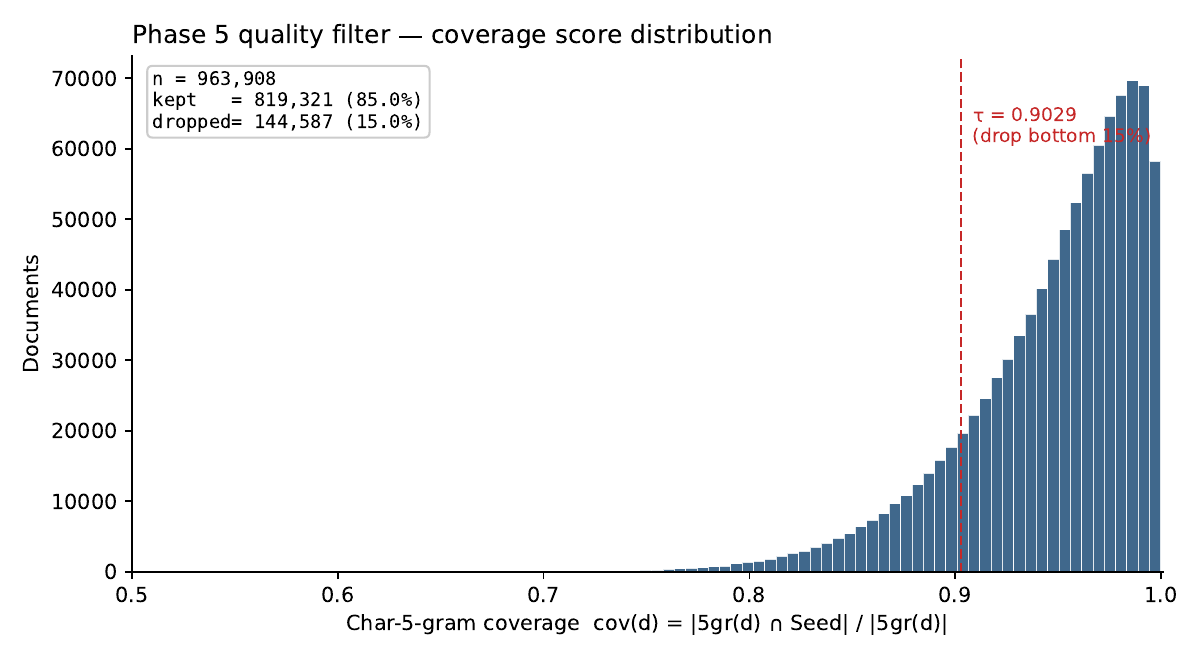}
  \caption{Char-5-gram coverage distribution (Phase 5) with $\tau = 0.9029$ drop threshold.}
  \label{fig:coverage}
\end{figure}

\section{Analysis}
\label{sec:analysis}

\paragraph{Dialect coverage.} GlotLID v3 identifies zero Maay Maay (\texttt{ymm\_Latn}) documents across the entire 1{,}077{,}804-document LID-verified corpus. This is likely a combination of Maay Maay being a primarily oral language with limited web presence, upstream sources filtering out non-standard dialectal forms, and GlotLID undercounting Maay on news-register text. We flag SomaliWeb v1 as \textbf{Standard Somali only}; v2 will source Maay Maay separately.

\paragraph{Source composition.}
\begin{table}[h]
\centering
\small
\begin{tabular}{lrr}
\toprule
Source & Kept docs & Fraction \\
\midrule
HPLT v2     & 582{,}257 & 71.07\% \\
CC100       & 233{,}394 & 28.49\% \\
Wikipedia   &   3{,}671 &  0.45\% \\
\bottomrule
\end{tabular}
\end{table}

\paragraph{Qualitative audit.} From a random sample of 20 release documents, all 20 were judged by a native Somali speaker as recognizable, well-formed Somali text suitable for pretraining. Full rubric-based audit deferred to a camera-ready appendix.

\section{Limitations}
\label{sec:limits}

\begin{enumerate}[leftmargin=1.5em,itemsep=2pt,topsep=2pt]
  \item \textbf{No downstream language-model evaluation in v1.} Tokenizer fertility is a proxy for downstream gains, not a substitute. We claim representational-compression wins at the tokenizer level only. Training small language models on this corpus and on phase-removed ablations, and comparing held-out perplexity, is the cleaner comparison and is the headline contribution planned for v2.
  \item \textbf{Quality filter is heuristic.} Character-5-gram coverage against a Wikipedia seed is interpretable and label-free, but it is not validated against human quality judgments at scale. It also downranks news heavy in Somali-language proper nouns. A trained classifier bootstrapped on SomaliWeb v1, plus correlation with manual rubric scoring, is planned for v2.
  \item \textbf{Small LID test set.} 200 rows total / 40 per class. We report 95\% bootstrap CIs but the intervals are wide. Single annotator (the first author). Expanded multi-annotator test set with Cohen's $\kappa$ is planned for v2.
  \item \textbf{No baseline-pipeline comparison.} We measure quality defects HPLT v2 carries, but we do not directly benchmark our pipeline against CCNet \citep{wenzek2020ccnet} or alternative dedup + filter stacks on the same Somali input. v2 will add this comparison.
  \item \textbf{Standard Somali only.} No Maay Maay coverage; pipeline would need dialect-aware LID adjustments.
  \item \textbf{No Somali-aware PII scrub.} Empirical scan: $\sim$7.9\% of release documents contain at least one email-shaped string. Presidio does not cover Somali. Consumer-facing downstream uses must apply additional PII filtering.
  \item \textbf{Source coverage window.} 2019--2024, inherited from HPLT v2 and CC100.
  \item \textbf{Inherited biases.} The Somali internet skews diaspora / news / politics / religion. Under-represented registers: conversational speech, technical writing, long-form fiction.
\end{enumerate}

\section{Conclusion}
\label{sec:conclusion}

SomaliWeb v1 is the first versioned, documented Somali pretraining corpus released with a companion tokenizer and language-identification benchmark. We release the corpus (819{,}322 documents, $\sim$303M tokens), a matched BPE-16K tokenizer (40.2\% more efficient than \texttt{cl100k\_base} on FLORES-200-so at the tokenizer-fertility level), and the first public per-class Somali LID benchmark across three production tools. Our measurements surface three concrete quality defects in the widely-used HPLT v2 ``cleaned'' Somali release. We frame the paper as an audit of an artifact rather than a downstream-modeling claim; v2 will add language-model perplexity comparisons across phase-removal ablations, an expanded multi-annotator LID test set, and a baseline-pipeline comparison against CCNet on the same Somali input.

\section*{Ethical Considerations}

\paragraph{Licensing.} Corpus inherits the most restrictive upstream license. Somali Wikipedia is CC-BY-SA 4.0; HPLT v2 is CC0; CC100 inherits CommonCrawl ToS. We release SomaliWeb v1 as CC-BY-SA 4.0 out of caution for the Wikipedia contribution.

\paragraph{Consent and attribution.} All three upstream sources are public and aggregated under licenses permitting redistribution. The dataset card attributes each source explicitly.

\paragraph{Potentially harmful content.} The Somali internet includes religious, political, and news content that may express views objectionable to some readers. We apply no content filtering; downstream users should apply their own.

\paragraph{PII.} See Limitations: $\sim$7.9\% of documents contain email-shaped strings. Downstream consumer-facing applications must perform additional PII scrubbing.

\paragraph{Dual use.} The corpus is suitable for pretraining, tokenizer training, and linguistic research; it is also usable for surveillance or disinformation applications. We cannot prevent such use but encourage users to consult the Responsible AI licensing literature.

\section*{Acknowledgments}

We thank the Somali Wikipedia community for maintaining the cleanest Somali text on the web. We thank the HPLT, CC100, FLORES-200, and GlotLID teams for releasing the upstream resources this work builds on. We thank Hugging Face for hosting the released corpus.

\bibliographystyle{plainnat}
\bibliography{refs}

\appendix

\section{Data Statement \citep{bender2018datastatements}}
\label{app:datastatement}

\textbf{A.1 Curation rationale.} SomaliWeb v1 is a pretraining corpus intended for language-model and tokenizer training on Standard Somali. Documents were selected by aggregating three upstream sources and applying a six-stage deduplication + cleaning + LID + quality filter.

\textbf{A.2 Language variety.} Standard Somali (ISO 639-3 \texttt{som}, Latin script). No Maay Maay (\texttt{ymm\_Latn}) coverage in v1.

\textbf{A.3 Speaker demographics.} Unknown. Upstream sources are web crawls; original authorship demographics are not recoverable.

\textbf{A.4 Annotator demographics.} The LID test set was annotated by a single Somali-speaking author. Camera-ready will add a second annotator and report inter-annotator agreement.

\textbf{A.5 Speech situation.} Asynchronous written text from news, blog, forum, and encyclopedic sources. No audio; no conversational register.

\textbf{A.6 Text characteristics.} Mean document length (after pipeline): $\approx$285 whitespace words; median $\approx$150. Genre distribution inherited from sources: $\approx$60\% news / blog (HPLT + CC100), $\approx$40\% CommonCrawl general web, $<$1\% encyclopedic (Wikipedia).

\textbf{A.7 Recording quality.} Digital-native text; no OCR.

\textbf{A.8 Other.} Source partition disclosed (HPLT 71.07\%, CC100 28.49\%, Wikipedia 0.45\%); release date 2026-04-26.

\section{Reproducibility checklist}
\label{app:repro}

\begin{itemize}[leftmargin=1.5em,itemsep=2pt,topsep=2pt]
\item Code released: \url{https://github.com/khaledyusuf44/somali-corpus} (MIT licensed)
\item Seeds fixed: \texttt{seed=0} throughout; \texttt{DetectorFactory.seed=0} for langdetect.
\item Package versions pinned: \texttt{requirements.txt} with exact specifiers.
\item Hardware specified: MacBook Pro M4 Pro / 24 GB unified memory.
\item Wall-clock budget reported per phase (\S\ref{sec:setup}).
\item Intermediate statistics released: \texttt{reports/*.json} + \texttt{reports/*.md}.
\item Dataset hash: SHA-256 of train + validation JSONL files recorded in \texttt{data/release/SHASUMS}.
\item Tokenizer hash: SHA-256 of \texttt{tokenizer\_somaliweb.json} recorded.
\item Configuration: \texttt{configs/pipeline.yaml} is the single source of truth for all knobs.
\item License: CC-BY-SA 4.0 (corpus); MIT (code); CC-BY 4.0 (paper).
\end{itemize}

\end{document}